\documentclass{article}

\usepackage[utf8]{inputenc}
\usepackage{mlspconf}
\usepackage{amsmath,amsthm,amssymb,amsfonts}
\usepackage{graphicx}
\usepackage{url}

\def\K{\mathbf{K}}

\def\E{\mathbb{E}}

\def\N{\mathcal{N}}

\def\bl{\boldsymbol\ell}

\def\y{\mathbf{y}}

\def\u{\mathbf{u}}
\def\uf{\mathbf{u}_{\mathbf{f}}}
\def\us{\mathbf{u}_\sigma}
\def\bs{{\boldsymbol\sigma}}
\def\z{\mathbf{z}}

\def\f{\mathbf{f}}

\def\x{\mathbf{x}}

\def\0{\mathbf{0}}
\def\cov{\mathrm{cov}}

\def\diag{\mathrm{diag}}

\def\R{\mathbb{R}}
\def\GP{\mathcal{GP}}

\DeclareMathOperator{\vect}{vec}

\copyrightnotice{978-1-5386-5477-4/18/\$31.00 {\copyright} 2018 European Union}
\toappear{2018 IEEE International Workshop on Machine Learning for Signal Processing, Sept.\ 17--20, 2018, Aalborg, Denmark}

\title{LEARNING STOCHASTIC DIFFERENTIAL EQUATIONS WITH GAUSSIAN PROCESSES WITHOUT GRADIENT MATCHING}
\name{Cagatay Yildiz$^{\star}$ ~~ Markus Heinonen$^{\star \dagger}$ ~~ Jukka Intosalmi$^{\star}$ ~~ Henrik Mannerström$^{\star}$ ~~ Harri Lähdesmäki$^{\star}$}
\address{$^{\star}$Dept.\ of CS, Aalto University, Finland \qquad  
    $^{\dagger}$Helsinki Inst.\ of Information Technology HIIT, Finland}
\begin{document}

\maketitle

\begin{abstract}
We introduce a novel paradigm for learning non-parametric drift and diffusion functions for stochastic differential equation (SDE). The proposed model learns to simulate path distributions that match observations with non-uniform time increments and arbitrary sparseness, which is in contrast with gradient matching that does not optimize simulated responses. We formulate sensitivity equations for learning and demonstrate that our general stochastic distribution optimisation leads to robust and efficient learning of SDE systems. 
\end{abstract}

%Gaussian process drift and diffusion functions of stochastic differential equation (SDE) models from observations. Existing approaches assume 

%Nonparametric stochastic differential equation (SDE) models identify the underlying system dynamics using gradient matching approximation that omit the 

%In stochastic differential equation (SDE) the underlying system dynamics are predominantly 
%In conventional ODE modelling coefficients of an equation driving the system state forward in time are estimated. However, for many complex systems it is practically impossible to determine the equations or interactions governing the underlying dynamics. In these settings, parametric ODE model cannot be formulated. Here, we overcome this issue by introducing a novel paradigm of nonparametric ODE modeling that can learn the underlying dynamics of arbitrary continuous-time systems without prior knowledge. We propose to learn non-linear, unknown differential functions from state observations using Gaussian process vector fields within the exact ODE formalism. We demonstrate the model's capabilities to infer dynamics from sparse data and to simulate the system forward into future.

\begin{keywords}
Stochastic differential equations, Gaussian processes
\end{keywords}

%Deep Gaussian processes (DGPs) are multi-layer generalisations of GPs, but inference in these models has proved challenging. 
%Existing approaches to inference in DGP models assume approximate posteriors that force independence between the layers, and do not work well in practice. 
%We present a doubly stochastic variational inference algorithm, which does not force independence between layers.
%With our method of inference we demonstrate that a DGP model can be used effectively on data ranging in size from hundreds to a billion points. 
%We provide strong empirical evidence that our inference scheme for DGPs works well in practice in both classification and regression. 

\section{Introduction}
\label{sec:intro}

Dynamical systems modeling is a cornerstone of experimental sciences. Modelers attempt to capture the dynamical behavior of a stochastic system or a phenomenon in order to improve its understanding and make predictions about its future state. Stochastic differential equations (SDEs) are an effective formalism for modelling systems with underlying stochastic dynamics, with wide range of applications \cite{friedrich2011}. The key problem in SDE's is estimation of the underlying deterministic driving function, and the stochastic diffusion component.

We consider the dynamics of a multivariate system governed by Markov process $\x_t$ described by an SDE
\begin{align} 
d\x_t = \f(\x_t)dt + \sigma(\x_t) dW_t   \label{eq:dx}
\end{align}
where $\x_t \in \R^D$ is the state vector of a $D$-dimensional dynamical system at continuous time $t \in \R$, $\f( \x ) \in \R^D$ is a deterministic state evolution, $\sigma(\x) \in \R$ is a scalar magnitude of the stochastic multivariate Wiener process $W_t \in \R^D$. The Wiener process has zero initial state $W_0 = \0$, and the independent increments $W_{t+s} - W_{t} \sim \N(\0, s I)$ follow a Gaussian with standard deviation $\sqrt{s}$. 

The SDE system \eqref{eq:dx} transforms states $\x_t$ forward in continuous time by the deterministic \emph{drift} component $\f$, while the $\sigma$ is the magnitude of the random Brownian \emph{diffusion} $W_t$ that scatters the state $\x_t$ with random fluctuations. The state solutions of SDE are given by the It\^o integral \cite{oksendal2014}
\begin{align}
\x_t &= \x_0 + \int_0^t \f(\x_\tau) d\tau + \int_0^t \sigma(\x_\tau) d W_\tau, \label{eq:ito}
\end{align}
where we integrate the system state from an initial state $\x_0$ for time $t$ forward, and where $\tau$ is an auxiliary time variable. The only non-deterministic part of the solution \eqref{eq:ito} is the Brownian motion $W_\tau$, whose random realisations generate path realisations $\x_{0 \ldots t}$ that induce state distributions $p(\x | t; \f, \sigma)$ at time $t$ given the drift $\f$ and diffusion $\sigma$.
%SDEs describe continuous-time systems with deterministic state evolution with Brownian diffusion that accounts for random fluctuations in the state $\x$. 
SDEs produce continuous, but non-smooth trajectories $\x_{0 \ldots t}$ over time due to the non-differentiable Brownian motion. 

%We often assume that the state distribution $p_0(\x) = \delta(\x - \x_0)$ is a Dirac delta function at time zero for some start state $\x_0$.
%An SDE implies state distributions $p(\x | t ; \f, \sigma)$ that are transformed forward in time 

We assume that both $\f(\cdot)$ and $\sigma(\cdot)$ are \emph{completely unknown} and we only observe one or several multivariate time series $Y = (\y_1, \ldots, \y_N)^T \in \R^{N \times D}$ obtained from noisy observations at observation times $T = (t_1, \ldots, t_N) \in \R^N$,
\begin{equation}
\y_t = \x_t + \varepsilon_t,
\end{equation}
where $\varepsilon_t \sim \N(\0, \Omega)$ follows a stationary time-invariant zero-mean multivariate Gaussian distribution with diagonal noise variances $\Omega = \diag( \omega_1^2, \ldots, \omega_D^2)$; and the latent states $\x_t \sim p(\x | t; \f, \sigma)$ follow the state distribution. %The observation time points do not need to be equally spaced. 
The goal of SDE modelling is to learn the drift $\f$ and diffusion $\sigma$ functions such that the process $\x_t$ matches data $\y_t$.

There is considerable amount of literature on inferring SDEs that have a pre-defined parametric drift or diffusion functions for specific applications \cite{friedrich2011}. There has also been interest on estimating non-parametric SDE drift and diffusion functions from data using the general Bayesian formalism \cite{ruttor2013,garcia2017}. With linear drift approximations the state distribution turns out to be a Gaussian, which can be solved with variational smoothing algorithm \cite{archambeau2007} or by variational mean field approximation \cite{vrettas2015}. Non-linear drifts and diffusions are predominantly modelled with Gaussian processes \cite{ruttor2013,batz2017,garcia2017}, which are a family of Bayesian kernel methods \cite{rasmussen2006}. For such models the state distributions are intractable, and hence these methods resort to using a family of gradient matching approximations \cite{Varah1982,Ellner2002,ramsay2007}, where the drift is estimated to match the empirical gradients of data, $\f(\y_i) \approx (\y_{i+1} - \y_i) \Delta t_i$, and the diffusion relates to the residual of the approximation \cite{ruttor2013,batz2017,garcia2017}. The gradient matching is only applicable to dense observations over time, while additional linearisation \cite{ruttor2013,batz2017} is necessary to model sparse observations. Non-parametric estimation of diffusion only applies to dense data \cite{batz2017,garcia2017}.

\begin{figure*}[t]
\centering \includegraphics[width=\textwidth]{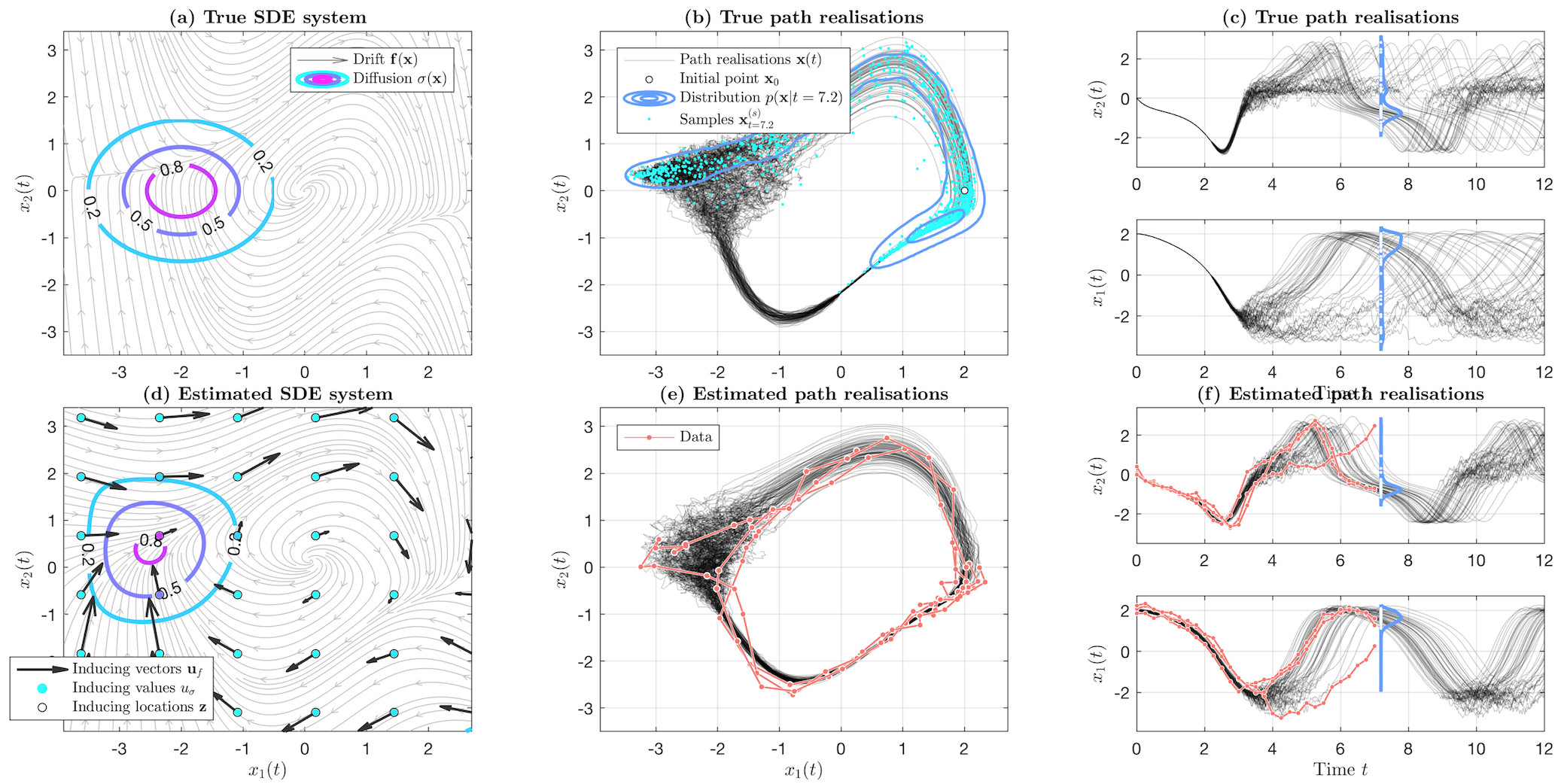}
\caption{\textbf{(a)} A Van der Pol oscillator with local diffusion and drift, \textbf{(b-c)} path samples and distribution, \textbf{(d)} the estimated GP system, \textbf{(e)} three noisy input trajectories for training, \textbf{(e-f)} the estimated path samples that match the true samples.}
\label{fig:vdp}
\end{figure*}

%% papas12: 1D true likelihood but its always gaussian
% archambeau07: true likelihood with linear approximation
% vrettas15: true likelihood with linear approximation

% ruttor13: dense data, GM direct, linear for sparse. no diffusion
% batz17: GM for dense, EM for sparse. non-linear diffusion for dense only with GM
% garcia17: dense data, GM + variational laplace. non-linear diffusion for dense only with GM

%Similarly, several methods have been published on inferring non-parametric drift and diffusion functions with Bayesian or Gaussian process formalisms [REFS]. Previous methods for non-parametric SDEs use the family of gradient matching approximations where the drift is estimated to match the empirical gradients of data, $\f(\y_i) \approx (\y_{i+1} - \y_i) \Delta t_i$, and the diffusion relates to the residual of the approximation \cite{Varah1982,Ellner2002,ramsay2007,garcia2017}. The gradient matching approximations are susceptible to bias due to not solving the data $\y$ against forward simulated responses $\x_t$ from the SDE.

%In this paper we propose to infer non-parametric drift and diffusion functions with Gaussian processes from observations $\y_i$ against the true forward-simulated expected likelihood of the state distributions $p(\x | t, \f, \sigma)$,
In this paper we propose to infer non-parametric drift and diffusion functions with Gaussian processes for arbitrary sparse or dense data. We learn the underlying system to induce state distributions with high expected likelihood,
%against the true forward-simulated expected likelihood of the state distributions $p(\x | t, \f, \sigma)$,
\begin{align}
p(Y | \f, \sigma, \Omega ) &= \prod_{i=1}^N \E_{p(\x | t_i ; \f, \sigma)} [\N(\y_i | \x, \Omega)]  \label{eq:lik}. 
%p(Y | \f, \sigma ) &= \E_{p(\x(0:t_N) | \f, \sigma)}  \N(\y_i | \x(t_i), \Omega)  \label{eq:lik}. 
%\log p(Y) &= \sum_{i=1}^N \log \int p(\y_i | \x) p(\x | t_i) d \x \\
%      &= \prod_{i=1}^N \int \N(\y_i | \x, \sigma_n^2 I) p_{t_i}(\x) d\x \\
%      &\approx \prod_{i=1}^N \sum_{s=1}^M \N(\y_i | \x^{(s)}, \sigma_n^2 I), \quad \x^{(s)} \sim p_{t_i} \\
%      & \hspace{-9mm} \approx \prod_{i=1}^N \frac{1}{M} \sum_{s=1}^M \N(\y_i | \x^{(s)}, \Omega), \,\,\, \x^{(s)} \sim p(\x | t_i ; \f, \sigma) 
\end{align}
The expected likelihood is generally intractable. In contrast to earlier works we do not use gradient matching or other approximative models, but instead we directly tackle and optimize the SDE system against the true likelihood \eqref{eq:lik} by performing full forward simulation. We propose an unbiased stochastic Monte Carlo approximation for the likelihood, for which we derive efficient, tractable gradients. Our approach places no restrictions on the spacing or sparsity of the observations. Our model is denoted as \textsc{npSDE}, and the implementation is publicly available in \url{http://www.github.com/cagatayyildiz/npde}.

\section{Inducing Gaussian process SDE model}

In this section we model both the drift $\f(\x)$ and diffusion $\sigma(\x)$ as Gaussian processes with a general inducing point parameterisation. The drift function defines a \emph{vector field} $\f : \R^D \rightarrow \R^D$, that is, an assignment of a $D$-dimensional gradient vector $\f(\x) \in \R^D$ to every $D$-dimensional state $\x \in \R^D$. We assume that drift does not depend on time. The diffusion function $\sigma(\x) \in \R$ is a standard scalar function. We model both functions as Gaussian processes (GP), which are flexible Bayesian non-linear and non-parametric models \cite{rasmussen2006}.

\subsection{Drift Gaussian process}

The inducing point parameterisation for the drift GP was originally proposed in the context of ordinary differential equation systems \cite{heinonen18a}, which we review here. We assume a zero-mean vector-valued GP prior on the drift function
\begin{align}
\f(\x) &\sim \GP( \0, K_\f(\x,\x') ),
\end{align}
which defines \emph{a priori} distribution over drift values $\f(\x)$ whose mean and covariance are
\begin{align}
\E[ \f(\x)] &= \0 \\
\cov[ \f(\x), \f(\x')] &= K_\f(\x,\x')  \,\,\,\, \in \R^{D \times D},
\end{align}
where the kernel $K_\f(\x,\x')$ is matrix-valued \cite{alvarez2012}. A GP prior defines that for any collection of states $X = (\x_1, \ldots, \x_N)^T \in \R^{N \times D}$, the drift values $F = ( \f(\x_1), \ldots, \f(\x_N))^T \in \R^{N \times D}$ follow a matrix-valued normal \cite{alvarez2012},
\begin{align}
p(F) &= \N(\vect F | \0, \K_\f(X,X) ),
\end{align}
where $\K_\f(X,X) = (K_\f(\x_i, \x_j))_{i,j=1}^N \in \R^{ND \times ND}$ is a block matrix of matrix-valued kernels $K_\f(\x_i,\x_j)$. The key property of Gaussian processes is that they encode functions where similar states $\x,\x'$ induce similar drifts $\f(\x),\f(\x')$, and where the state similarity is defined by the kernels $K_\f(\x,\x')$. Several families of rich matrix-valued kernels exist
%macedo2008
\cite{wahlstrom2013,alvarez2012,micchelli2005}. In this work we opt for the family of decomposable kernels $K(\x,\x') = k(\x,\x') \cdot A$, where $k(\x,\x')$ is a Gaussian base kernel
\begin{align}
k(\x,\x') &= \sigma_\f^2 \exp \left( - \frac{1}{2} \sum_{d=1}^D \frac{(x_d-x_d')^2}{\ell_{\f d}^2} \right) \label{eq:gauss}
\end{align}
with drift variance $\sigma_\f^2$, and dimension-specific lengthscales $\ell_{\f 1}, \ldots, \ell_{\f D}$ that determine the smoothness of drift field, and $A \in \R^{D \times D}$ is a PSD dependency matrix between dimensions. In practise global dependency structures are often unavailable, and the diagonal structure $A = I_D$ is then chosen.

%We note that more complex kernels can also be considered given prior information of the underlying system characteristics. The divergence-free matrix-valued kernel induces vector fields that have zero divergence \citep{wahlstrom2013,solin2015}. Intuitively, these vector fields do not have sinks or sources, and every state always finally returns to itself after sufficient amount of time. Similarly, curl-free kernels induce curl-free vector fields that can contain sources or sinks, that is, trajectories can accelerate or decelerate. For theoretical treatment of vector field kernels, see \citep{narcowich1994generalized,bhatia2013,fuselier2017}. Non-stationary vector fields can be modeled with input-dependent lengthscales \citep{heinonen2016}, while spectral kernels can represent stationary \citep{wilson2013} or non-stationary \citep{remes2017} recurring patterns in the differential function.

In standard GP regression we would obtain posterior of the drift by conditioning the GP prior with data \cite{rasmussen2006}. In SDE models the conditional $\f(\x) | Y$ is intractable due to the integral mapping \eqref{eq:ito} between observations $\y_i$ and drifts $\f(\x)$. Instead, we augment the Gaussian process with a set of $M$ \emph{inducing vectors} $\uf \in \R^D$ at locations $\z \in \R^D$, such that $\f(\z) = \uf$ \cite{quinonero2005}. We interpolate drift from inducing points as
\begin{align}
\f(\x) &\triangleq \K(\x, Z) \K(Z,Z)^{-1} \uf, \label{eq:f}
\end{align}
which supports the drift with \emph{inducing locations} $Z = (\z_1, \ldots, \z_M)$ and \emph{inducing vectors} $U_\f = (\u_{\f 1}, \ldots, \u_{\f M})$, and where $\uf = \vect U_\f$. This corresponds to a vector-valued kernel function \cite{alvarez2012}, or to a multi-task Gaussian process posterior mean \cite{rasmussen2006}. Due to universality of the Gaussian kernel \cite{shawe2004kernel}, we can represent arbitrary drifts with sufficient inducing points.

\subsection{Diffusion Gaussian process}

We represent diffusion $\sigma(\x)$ as another inducing point GP, similarly to drift. Diffusion is a scalar function that uses a scalar kernel. The diffusion has a zero-mean GP prior
\begin{align}
\sigma(\x) &\sim \GP( 0, k_{\sigma}(\x,\x') )
\end{align}
that defines covariance $\cov[ \sigma(\x), \sigma(\x')] = k_\sigma(\x,\x') \in \R$ with a Gaussian kernel of form \eqref{eq:gauss}, but with diffusion variance $\sigma_\sigma^2$ and lengthscales $\{ \ell_{\sigma d}\}$. % with $k(\x,\x') = \sigma_\sigma^2 \exp \left( - 1/2 \sum_{d=1}^D (x_d-x_d')^2 / \ell_{\sigma d}^2 \right)$. 
Diffusion values $\bs = (\sigma(\x_1), \ldots, \sigma(\x_N))^T \in \R^N$ at states $X$ then follow a prior 
\begin{align}
p(\bs) &= \N( \bs | \0, K_\sigma(X,X) ),
\end{align}
where $K_\sigma(X,X) = ( k_\sigma(\x_i,\x_j) )_{i,j=1}^N \in \R^{N \times N}$. We interpolate the diffusion from $M$ inducing locations $Z$ with inducing values $\us = (u_{\sigma 1}, \ldots, u_{\sigma M})^T \in \R^M$,
%values $u_\sigma \in \R$ at locations $\z$, such that $\sigma(\z) = u_\sigma$, and interpolate the diffusion function between inducing points as
\begin{align}
\sigma(\x) &\triangleq K_\sigma(\x, Z) K_\sigma(Z,Z)^{-1} \us, \label{eq:sigma}
\end{align}
Drift and diffusion naturally share their inducing locations $Z$.

\subsection{Stochastic Monte Carlo inference}

The inducing SDE model is determined via the inducing locations $Z$, the inducing values $\uf$ and $\us$, the observation noise variance $\sigma_n^2$, and the kernel parameters $\sigma_\sigma,\{\ell_{\sigma d}\}$ and $\sigma_\f,\{\ell_{\f d}\}$ of the drift and diffusion kernels. 
%We assume uniform priors for the kernel parameters. If domain knowledge of them is available, they can be included in the Bayesian model in a straightforward manner. 
The posterior of the model combines the likelihood $p(Y | \f, \sigma, \Omega)$ of \eqref{eq:lik} and the independent priors $p(\uf)$ and $p(\us)$ using Bayes' theorem as
\begin{align}
    p(\uf, \us | Y) &\propto p(\uf, \us) p(Y | \uf, \us) \label{eq:post} \\
    &= p(\uf)  p(\us) \prod_{i=1}^N \E_{p(\x | t_i ; \f, \sigma)} [ \N(\y_i | \x, \Omega) ] \notag \\
%    &= \prod_{i=1}^N \E_{p(\x | t_i ; \uf, \us, Z)} [ p(\y_i | \x) ] \\
    & \hspace{-16mm} \approx \N( \uf | \0, \K_\f(Z,Z) ) \N( \us | \0, K_\sigma(Z,Z) ) \label{eq:post_stoc} \\
    & \hspace{-12mm} \times \prod_{i=1}^N \frac{1}{N_s} \sum_{s=1}^{N_s} \N(\y_i | \x_i^{(s)}, \Omega), \,\,\, \x^{(s)} \sim p(\x_{0 \ldots t} |  \uf, \us, Z) \notag
\end{align}
where the time-dependent state distribution $p(\x | t ; \f, \sigma) \equiv p(\x | t ; \uf, \us, Z)$ now depends on the inducing parameters. We propose to approximate the true expected likelihood with an unbiased stochastic Monte Carlo averaging, since we can draw path samples $\x_t^{(s)}$ from the state distribution $p(\x_{0 \ldots t} | \uf,\us,Z)$ by sampling the Brownian motion path $W_t^{(s)}$. The stochastic likelihood estimate with $N_s$ samples turns out to be a kernel density estimator with Gaussian bases. 

We draw the sample paths using Euler-Maruyama(EM) method  for approximating the solution of an SDE \eqref{eq:ito} \cite{oksendal2014}:
\begin{align}
    \x_{i+1}^{(s)} &= \x_{i}^{(s)} + \f(\x_i^{(s)}) \Delta t + \sigma(\x_i^{(s)}) \Delta W_i^{(s)}, \label{eq:em}
\end{align}
where we discretise time into $N_T$ subintervals $t_0, t_1, \ldots, t_{N_T}$ of width $\Delta t = t_{N_T} / N_T$, and sample the Wiener coefficients as $\Delta W_i^{(s)} \sim \N(\0, \Delta t \cdot I)$ with standard deviation $\sqrt{\Delta t}$. We set $\x_0^{(s)}$ to the initial observation and use \eqref{eq:em} to compute state path $\x^{(s)} \equiv (\x_0^{(s)}, \x_1^{(s)}, \ldots, \x_{T_N}^{(s)})$. The number of time steps $N_T > N$ is often higher than the number of observed timepoints to achieve sufficient path resolution.

\begin{figure}[th]
    \centering
    \includegraphics[width=\columnwidth]{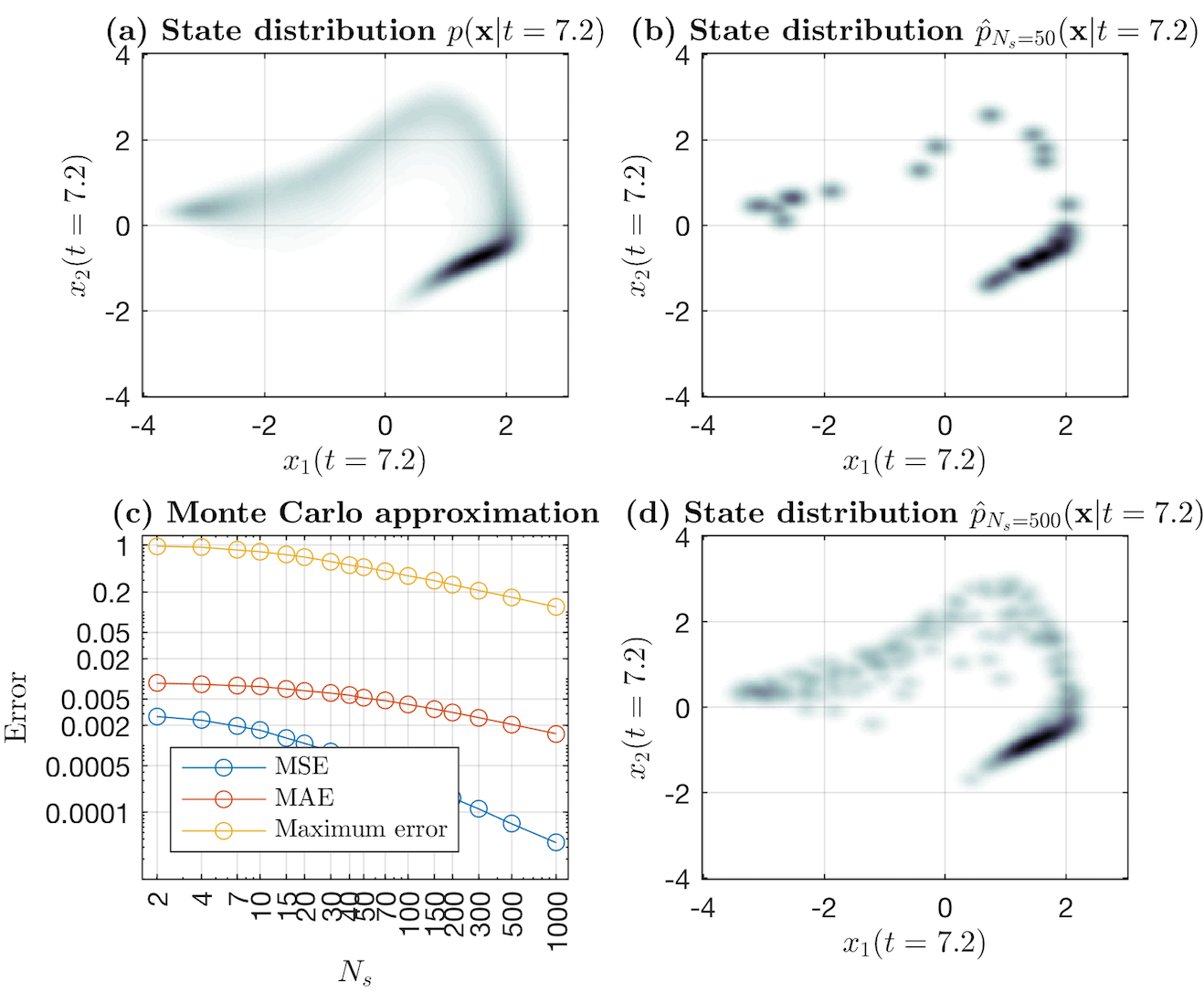}
    \caption{\textbf{(a)} True state distribution from the model in Figure \ref{fig:vdp}(a-b), \textbf{(b,d)} state distribution approximations with different number of path samples, \textbf{(c)} approximation errors.}
    \label{fig:mcerr}
\end{figure}

%The likelihood resembles a particle filter, or a kernel density estimator with a Gaussian kernel with noise bandwidths. The likelihood is intractable, but can be numerically approximated by drawing $M$ trajectory realisations using the Ito integral starting from state $\x_0$ at time $0$ (See Figure \ref{fig:vdp}b).

%The drift and diffusion functions $\f(\x)$ and $g(\x)$ imply a state distribution $p_t(\x)$ through the SDE forward simulation equation \eqref{eq:ito}. We do not have a tractable form of the state distribution $p_t$, but we can draw state trajectory $\x_t$ samples using equation \eqref{eq:ito} by sampling realisations of the stochastic Wiener coefficients $W_\tau$ through a discretisation method, such as Euler-Maruyama or stochastic Runge-Kutta approximations.

%To optimize the approximative posterior \eqref{eq:post_stoc} we first apply latent reparameterisations using Cholesky decompositions $\L_\f \L_\f^T = \K_\f(Z,Z)$ and $L_\sigma L_\sigma^T = K_\sigma(Z,Z)$,
%\begin{align}
%    \uf &= \L_\f \tuf, \qquad \us = L_\sigma \tus,
%\end{align}
%which projects the whitened latent parameters $\tuf$ and $\tus$ on to the smooth kernel manifold induced by the Cholesky decompositions \cite{kuss2005,anon2018}, with the whitened parameters having standard normal priors $\tuf \sim \N(\0,I)$ and $\tus \sim \N(\0,I)$. Optimizing the whitened parameter vectors is very efficient since they are fully decoupled from each other. 
%We refer the reader to the Appendix for detailed derivation of the whole whitening procedure. 

We find a maximum \emph{a posteriori} (MAP) estimates of $\uf,\us,\Omega$ by gradient ascent, while choosing lengthscales $\bl_\f,\bl_\sigma$ from a grid, keeping the inducing locations $Z$ fixed on a dense grid (See Figure \ref{fig:vdp}(d)) and setting $\sigma_\f=\sigma_\sigma=1$. In practise in $2D$ or $3D$ systems placing inducing locations on a grid is a robust choice, noting that they can be also optimised with increased computational complexity \cite{hensman2013}.

\subsection{Computing stochastic gradients}

%The key term to carry out the MAP gradient ascent optimization are the likelihoods $\N(\y_i | \x_i^{(s)},\Omega)$ within the log of the likelihood \eqref{eq:post_stoc} of the frozen state paths $\x_t^{(s)}$, which depends on the inducing vector parameters $\u \triangleq (\uf,\us)$;
%\begin{align}
%&\hspace{-5mm} \frac{\sum_{i=1}^N \log \frac{1}{N_s} \sum_{s=1}^{N_s} d  p(\y_i | \x_i^{(s)}, \Omega)}{d \u} \\
%&= \sum_{i=1}^N \log \frac{1}{N_s} \sum_{s=1}^{N_s} \frac{d \N(\y_i | \x_i^{(s)}, \Omega)}{d\x} \frac{d\x^{(s)}_i}{d \u}.
%\end{align}

%Markus, could this be simply:

The gradient of the expectation of the log-likelihood \eqref{eq:post_stoc} is
\begin{align}
&\hspace{-5mm} \frac{d}{d\u} \sum_{i=1}^N \log \frac{1}{N_s} \sum_{s=1}^{N_s} \N(\y_i | \x_i^{(s)}, \Omega) \\
&= \sum_{i=1}^N \frac{ \sum_{s=1}^{N_s} \frac{\partial \N(\y_i | \x_i^{(s)}, \Omega)}{\partial \x} \frac{d\x^{(s)}_i}{d \u}}{ \sum_{s=1}^{N_s} \N(\y_i | \x_i^{(s)}, \Omega)}, \label{eq:postgrad}
\end{align}
where the sample paths $\x_t^{(s)}$ are from equation \eqref{eq:em}. The last term $\frac{d \x_i^{(s)}}{d \u}$ is the cumulative derivative of the state $\x_i^{(s)}$ of sample $s$ at time $t_i$ against the parameters $\u \triangleq (\uf,\us)$. 
%The sample paths are in practise constructed by using the Euler-Maruyama method for approximating the solution of an SDE \eqref{eq:ito}:
%\begin{align}
%    \x_{i+1}^{(s)} &= \x_{i}^{(s)} + \f(\x_i^{(s)}) \Delta t + \sigma(\x_i^{(s)}) \Delta W_i^{(s)}, \label{eq:em}
%\end{align}
%where we discretise time into $N_T$ equal subintervals $0 = t_0, t_1, \ldots, t_{N_T}$ of width $\Delta t = t_{N_T} / N_T$, and sample the Wiener coefficients as $\Delta W_i^{(s)} \sim \N(\0, \Delta t \cdot I)$ with standard deviation $\sqrt{\Delta t}$. We set the initial state of sample $s$ to $\x_0^{(s)}$ and use the recursion \eqref{eq:em} to compute state path $\x^{(s)} \equiv \x_0^{(s)}, \x_1^{(s)}, \ldots, \x_{T_N}^{(s)}$. The number of time steps $N_T > N$ is in general much higher than the number of observed timepoints to achieve sufficient path resolution.
%
%The Euler-Maruyama iteration for sample paths $\x^{(s)}$ is an piecewise approximation of a deterministic ODE.
The gradients of the piecewise Euler-Maruyama paths $\x_i^{(s)}$ are:
\begin{align}
\frac{d\x_{i+1}^{(s)}}{d\u} &= \frac{d\x_{i}^{(s)}}{d\u} + \frac{d\f(\x_{i}^{(s)})}{d\u} \Delta t + \frac{d\sigma(\x_{i}^{(s)})}{d\u} \Delta W_i^{(s)} \notag \\
 &= \frac{d\x_{i}^{(s)}}{d\u} + \left( \frac{\partial \f(\x_i^{(s)})}{\partial \x}  \frac{d\x_i^{(s)}}{d\uf} + \frac{\partial \f(\x_i^{(s)})}{\partial \uf} \right) \Delta t \label{eq:grad} \\
 &\hspace{13mm}+ \left( \frac{\partial \sigma(\x_i^{(s)})}{\partial \x}  \frac{d\x_i^{(s)}}{d\us} + \frac{\partial \sigma(\x_i^{(s)})}{\partial \us} \right) \Delta W_i^{(s)}. \notag
\end{align}
The derivatives $\frac{d\x_i^{(s)}}{d\u}$ are constructed iteratively over time starting from $\frac{d\x_0^{(s)}}{d\u} = \0$ with a fixed initial state $\x_0^{(s)}$, and the four partial derivatives are gradients of the kernel functions \eqref{eq:f} and \eqref{eq:sigma} with respect to $\uf$ and $\us$, respectively:
\begin{align}
\frac{\partial \f(\x)}{\partial \x} &= \frac{\partial \K_\f(\x,Z)}{\partial \x} \K_\f(Z,Z)^{-1} \uf  \\
\frac{\partial \f(\x)}{\partial \uf} &= \K_\f(\x,Z) \K_\f(Z,Z)^{-1} \\
\frac{\partial \sigma(\x)}{\partial \x} &= \frac{\partial K_\sigma(\x,Z)}{\partial \x} K_\sigma(Z,Z)^{-1} \us  \\
\frac{\partial \sigma(\x)}{\partial \us} &= K_\sigma(\x,Z) K_\sigma(Z,Z)^{-1}.
\end{align}
These gradients are related to the sensitivity equations \cite{kokotovic1967,froechlich2017} derived for non-parametric ODEs previously \cite{heinonen18a}. The gradients \eqref{eq:grad} can be computed in practise together with the sample paths \eqref{eq:em} during the Euler-Maruyama iteration. The computation of the gradients has the same computational complexity as the numerical simulator. The iterative gradients are superior to finite difference approximation since we have exact formulation of the gradients, albeit of the approximal Euler-Maruyama paths, which can be solved to arbitrary numerical accuracy by tuning $\Delta t$ discretisation.

\section{Experiments}

In order to illustrate the performance of our model, we conduct several experiments on synthetic data as well as real-world data sets. In all experiments inducing vectors are initialized by gradient matching, and then fully optimised. We use EM method to simulate the state distributions and compute stochastic gradients over $N_s = 50$ samples. We use L-BFGS algorithm to compute the MAP estimates.

%We use Euler-Maruyama approximation with step size $\Delta t = 0.005$ to generate synthetic data. In double well and two dimensional oscillator experiments, we observe a noisy version of every 20'th and 50'th state, respectively. The observation noise is Gaussian distributed with standard deviation 0.1. In all experiments, inducing points are placed uniformly on a grid that covers the data, and inducing vectors are initialized by gradient matching. We use L-BFGS algorithm to compute the MAP estimate. Whenever we aim to infer a constant diffusion, we set the diffusion length-scale to $10^6$, which effectively results in constant estimates.

\subsection{Double well}

\begin{figure}[t]
    \centering
    \includegraphics[width=\columnwidth]{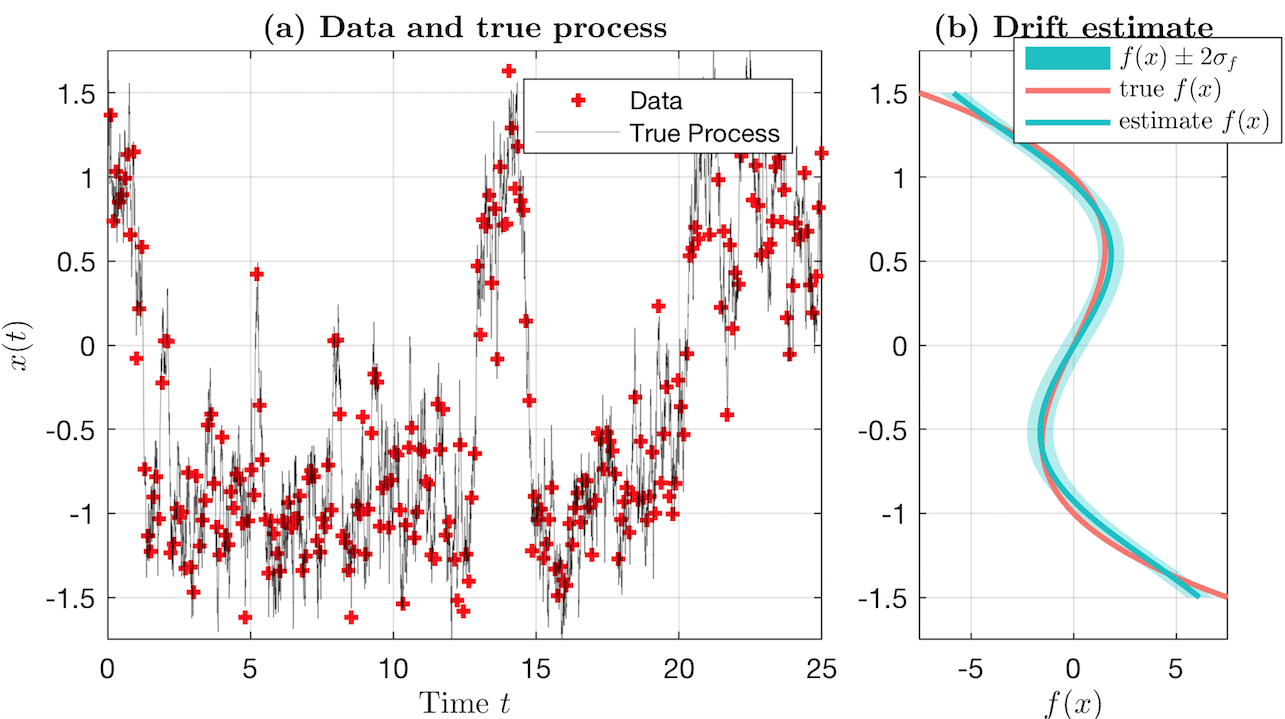}
    \caption{\textbf{(a)} Double well system, \textbf{(b)} estimated drift.}
    \label{fig:dw}
\end{figure}

We first consider the double well system where the drift is given by $f(x) = 4(x-x^3)$ and with constant diffusion $\sigma(x) = 1.5$. We generate $6$ random noisy input trajectories, each with $250$ observed data points. True dynamics and the observed data points are illustrated in Figure \ref{fig:dw}a. We fit our npSDE model with $M = 15$ inducing points located uniformly within $[-5,5]$. We accurately approximate the true drift (the right plot on Figure \ref{fig:dw}), and learn a diffusion estimate of $1.39$.

\begin{figure}[th]
    \centering
    \includegraphics[width=\columnwidth]{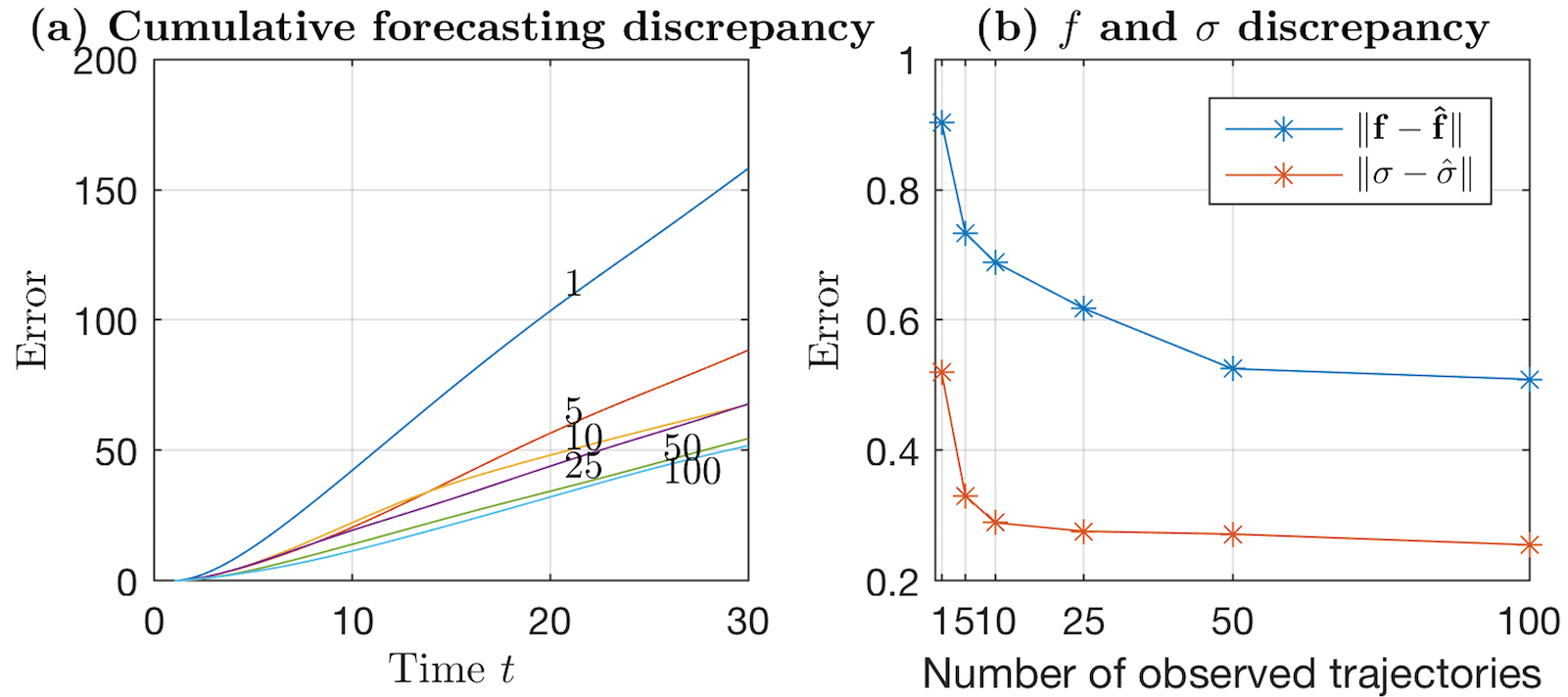}
    \caption{\textbf{(a)} Cumulative discrepancy in the state distributions over time and over number of observed trajectories in synthetic $2D$ model, \textbf{(b)} drift and diffusion estimation errors.}
    \label{fig:sm}
\end{figure}

\subsection{Simple oscillating dynamics}

Next, we investigate how the quality of fit changes by the amount of data used for training. We consider the $2D$ synthetic system in \cite{batz2017}, whose drift equations are given by $\f(\x)_1 = x_1(1-x_1^2-x_2^2)-x_2$ and $\f(\x)_2 = x_2(1-x_1^2-x_2^2)+x_1$, and the diffusion is $\sigma(\x) = 2\N(\x |[-1,-1], 0.5 I) + 0.3$. The state-dependent diffusion acts an hotspot of increased path scatter, and provides interesting and challenging dynamics to infer. We generate six data batches from the true dynamics using EM method with step size $\Delta t = 0.005$, and observe every 100'th state corrupted by a Gaussian noise with variance $\sigma_n^2=0.1^2$. The data batches contain 1, 5, 10, 25, 50 and 100 input trajectories, each having 25 data points. We repeat the experiments 50 times and report the average error.

The left plot in Figure~\ref{fig:sm} illustrates the cumulative discrepancy in the state distributions over time, and the right plot shows the error between the true and estimated drift/diffusion. Unsurprisingly, the discrepancy in both plots decrease when more training data is used. We also observe that the performance gain is insignificant after 50 input trajectories. 

\begin{figure*}[t]
    \centering
    \includegraphics[width=\textwidth]{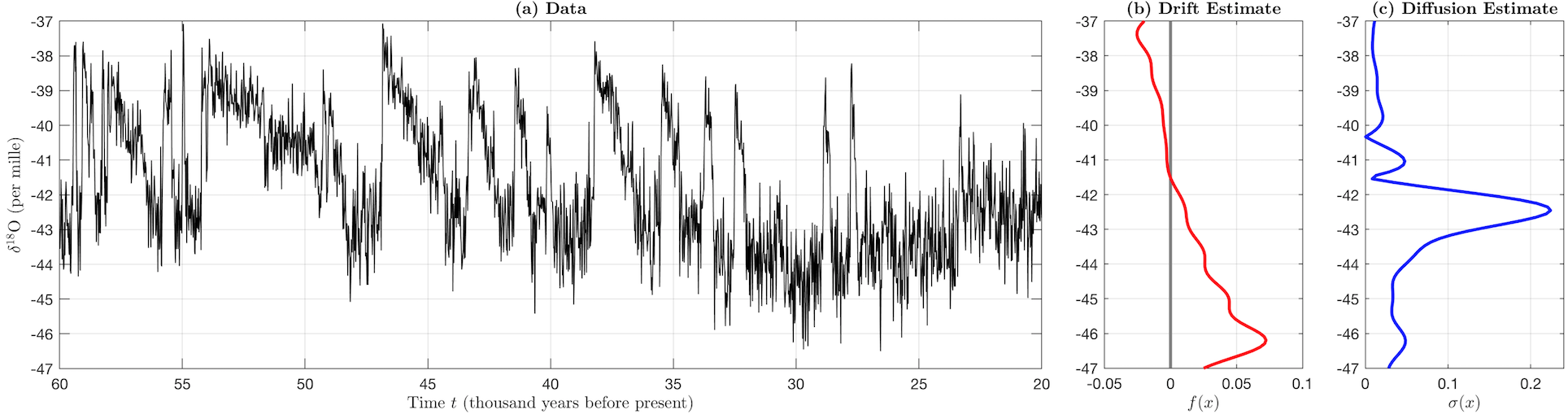}
    \caption{ \textbf{(a)} The ice core data, \textbf{(b)} estimated drift, \textbf{(c)} the highly state-dependent diffusion.}
    \label{fig:icecore}
\end{figure*}

\subsection{Ice core data}

As another showcase of our model, we consider the NGRIP ice core dataset \cite{andersen2004high}, which contains records of isotopic oxygen $\delta^{18}$O concetrations within clacial ice cores. The record is used to explore the climatic changes that date back to last glacial period. During that period, the North Atlantic region underwent abrupt climate changes known as Dansgaard-Oeschger (DO) events. The events are characterized by a sudden increase in the temperature followed by a gradual cooling phase. Following \cite{kwasniok2013analysis}, we consider $N=2000$ timepoints from the time span from 60000 years to 20000 years before present, where 16 DO events have been identified.
%\cite{rasmussen2014stratigraphic} 

Figure \ref{fig:icecore}(a) illustrates the highly variable data. We observe a repeating pattern of DO events: a sudden increase followed by a slower settlement phase. The panel \ref{fig:icecore}(b) indicates estimated drift that pushes the oxygen down until state $-41$ and up with small states, matching the data. Interestingly, diffusion at \ref{fig:icecore}(c) is highly peaked between $-42$ and $-43$, which has accurately identified the regime of DO events. The model has learned to explain the DO events with high diffusion.

%The middle and the rightmost panel shows the estimated drift and diffusion functions against time. The drift estimate is negative for states smaller than -41.5 and positive for the rest, which match the data. What's more interesting is the diffusion, which is relatively higher in states between -44 and -42. Because that interval is the same as the values observed  before the sudden jumps in the data, we interpret that the model fit \textit{explains} the jumps via diffusion.

\subsection{Human motion dataset}

\begin{figure*}[th]
    \centering
    \includegraphics[width=\textwidth]{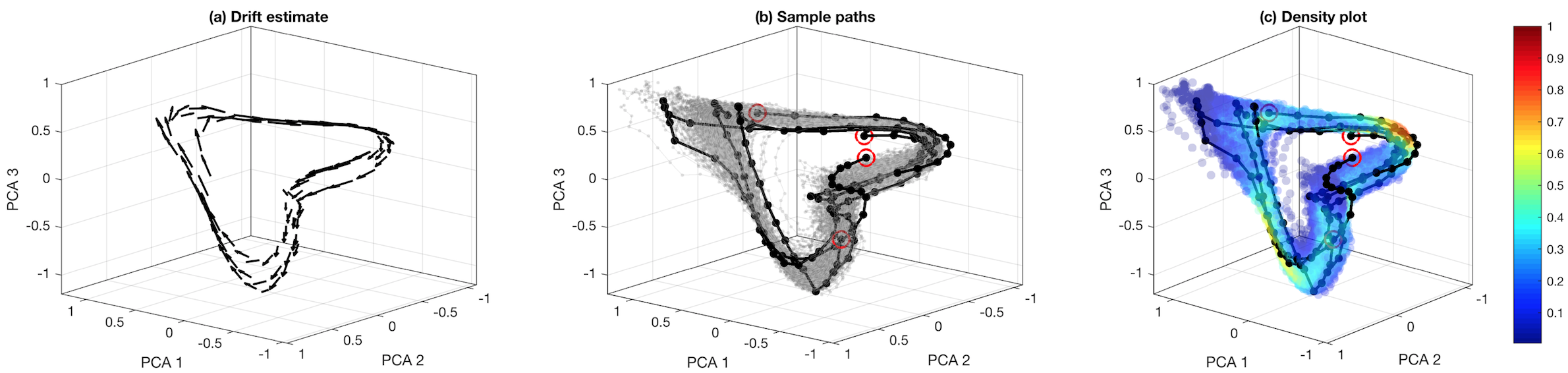}
    \caption{\textbf{(a)} Shared drift estimate learned from walking data of 4 subjects, \textbf{(b)} estimated sample paths, \textbf{(c)} density plots of the sample paths. The 4 observed trajectories are shown as black lines in \textbf{(b-c)}, with red circles denoting the initial state.}
    \label{fig:4walker}
\end{figure*}

We finally demonstrate our approach on human motion capture data. Our goal is twofold: to estimate a single drift function that captures the dynamics of the walking sequences of several people, and to explain the discrepancies among sequences via diffusion. The input trajectories are the same as in \cite{wang2008}: four walking sequences, each from a different person. We also follow the preprocessing method described in \cite{wang2008}, which results in a simplified skeleton that consists of 50-dimensional pose configurations. All records are mean centered and downsampled by a factor of two. 

Inference is performed in three dimensional space where the input sequences are projected using PCA. We place the inducing points on a $5\times 5\times 5$ grid and set the length-scale of both drift and diffusion process to 0.5. $3D$ data set, inferred drift fit and the density of the sample paths are visualized in Figure~\ref{fig:4walker}. We can conclude that our model is capable of inferring drift and diffusion functions that match arbitrary data.

%\begin{figure}
%    \centering
%    \includegraphics[width=0.9\columnwidth]{golf.jpg}
%    \caption{Golf swing data}
%    \label{fig:golf}
%\end{figure}

\section{DISCUSSION}

We propose an approach for learning non-parametric drift and diffusion functions of stochastic differential equation (SDE) systems such that the resulting simulated state distributions match data. Our approach can learn arbitrary dynamics due to the flexible inducing Gaussian process formulation. We propose a stochastic estimate of the simulated state distributions and an efficient system of computing their gradients. Our approach does not place any restrictions on the sparsity or denseness of the observations data. We leave learning of time-varying drifts and diffusions as interesting future work.

\paragraph*{Acknowledgements.} 
The data used in this project was obtained from \url{mocap.cs.cmu.edu}. The database was created with funding from NSF EIA-0196217. This work has been supported by the Academy of Finland Center of Excellence in Systems Immunology and Physiology, the Academy of Finland grants no.~260403, 299915, 275537, 311584.

\small

\bibliographystyle{IEEEbib}
\bibliography{refs}

\end{document}